\documentclass{article}

\pdfoutput=1

     \PassOptionsToPackage{numbers, compress}{natbib}

\usepackage[preprint]{neurips_2019}

\usepackage[utf8]{inputenc} 
\usepackage[T1]{fontenc}    
\usepackage{url}            
\usepackage{booktabs}       
\usepackage{amsfonts}       
\usepackage{nicefrac}       
\usepackage{microtype}      
\usepackage{adjustbox}
\usepackage{dcolumn}

\usepackage{epsfig}
\usepackage{graphicx}
\usepackage{amsmath}
\usepackage{amssymb}
\usepackage{xspace}
\usepackage{soul} 
\usepackage[dvipsnames]{xcolor}
\usepackage[linesnumbered,ruled]{algorithm2e}
\usepackage{wrapfig}
\usepackage{tikz}

\newcommand{\lpips}{\textsc{lpips}\xspace}
\newcommand{\ouralgo}{\textsc{e-lpips}\xspace}

\newcommand{\lpipsvgg}{\textsc{lpips-vgg}\xspace}
\newcommand{\lpipssqz}{\textsc{lpips-sqz}\xspace}

\DeclareMathOperator*{\argmax}{\text{arg max}}
\DeclareMathOperator*{\argmin}{\text{arg min}}

\newcommand{\E}{\mathbb{E}}

\newcommand{\lambdamin}{\lambda_{\text{min}}}
\newcommand{\lambdamax}{\lambda_{\text{max}}}

\newcommand{\del}[1]{\unskip}

\newcommand{\figdir}{figures_jpeg}
\newcommand{\h}{0mm}
\newcommand{\hh}{0mm}
\newcommand{\hhh}{0mm}
\newcommand{\hhhh}{0mm}

\newcommand{\geodesicRow}[3]{
\renewcommand{\hhhh}{1.98cm}
\rotatebox{90}{\rotatebox{-90}{\includegraphics[width=\hhhh,trim=#3,clip]{\figdir/geodesic/#1/L2/final_00.jpg}}\hfill
\rotatebox{-90}{\includegraphics[width=\hhhh,trim=#3,clip]{\figdir/geodesic/#1/L2/final_09.jpg}}}\hfill
\includegraphics[width=#2,trim=#3,clip]{\figdir/geodesic/#1/L2/whichrowMiddle.jpg}\hfill
\includegraphics[width=#2,trim=#3,clip]{\figdir/geodesic/#1/LPIPS-VGG/whichrowMiddle.jpg}\hfill
\includegraphics[width=#2,trim=#3,clip]{\figdir/geodesic/#1/ELPIPS/whichrowMiddle.jpg}\\
\makebox[\hhhh][r]{\raisebox{0.22cm}{\small Time Evol. $\rightarrow$}}\hfill
\includegraphics[width=#2,clip]{\figdir/geodesic/#1/L2/geodesic.jpg}\hfill
\includegraphics[width=#2,clip]{\figdir/geodesic/#1/LPIPS-VGG/geodesic.jpg}\hfill
\includegraphics[width=#2,clip]{\figdir/geodesic/#1/ELPIPS/geodesic.jpg}\\
} 

\newcommand{\figGeodesic}{
\begin{figure*}[t]
\begin{center}
\renewcommand{\h}{0.2825\linewidth}
\renewcommand{\hh}{1.5mm}
\makebox[2cm]{Input frames}\hfill
\makebox[\h]{$L_2$}\hfill
\makebox[\h]{\lpips}\hfill
\makebox[\h]{\ouralgo (ours)}\\
\geodesicRow{jp}{\h}{0 0 0 0}
\geodesicRow{pebbles}{\h}{0 32 0 32}
\caption{\label{fig:Geodesic}Discrete geodesics between Images A and B, computed in three metrics. Each column shows a single frame from the discrete geodesic, as well as the time evolution of the scanline indicated in red. The reader is encouraged to view the supplemental animations.}

\end{center}
\end{figure*}
}

\newcommand{\figContradictionAA}{
\begin{figure*}[t]
\begin{center}
\renewcommand{\h}{0.1625\linewidth}
\renewcommand{\hh}{-1mm}
\renewcommand{\hhh}{1mm}
\makebox[\h][c]{Source $a$}\hfill
\makebox[\h][c]{Anchor}\hfill
\makebox[\h][c]{Target $b$}\hfill
\parbox{\h}{\centering Attack on\\\lpipssqz}\hfill
\parbox{\h}{\centering Attack on\\\lpipsvgg}\hfill
\parbox{\h}{\centering Attack on\\\ouralgo (our)}\\
\includegraphics[width=\h]{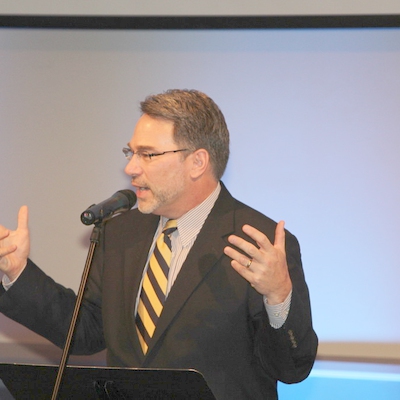}\hfill
\includegraphics[width=\h]{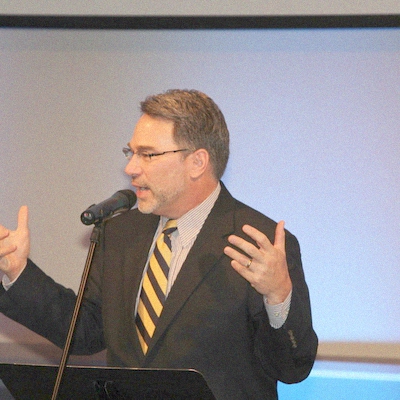}\hfill
\includegraphics[width=\h]{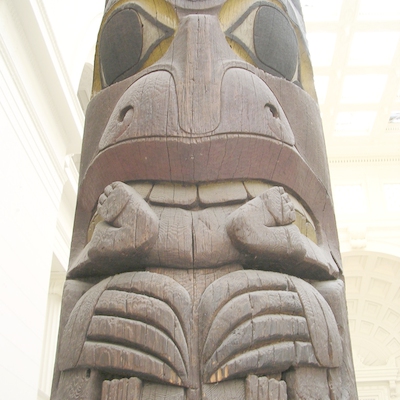}\hfill
\includegraphics[width=\h]{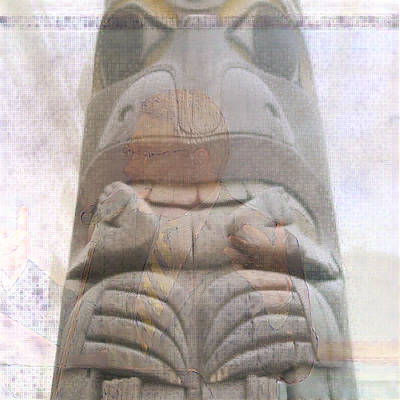}\hfill
\includegraphics[width=\h]{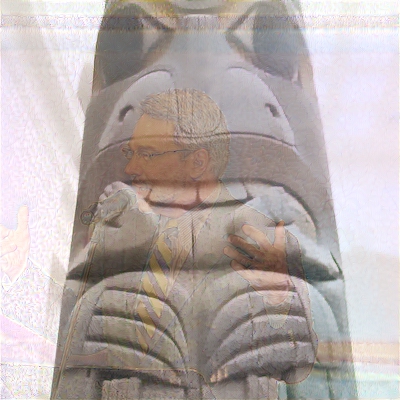}\hfill
\includegraphics[width=\h]{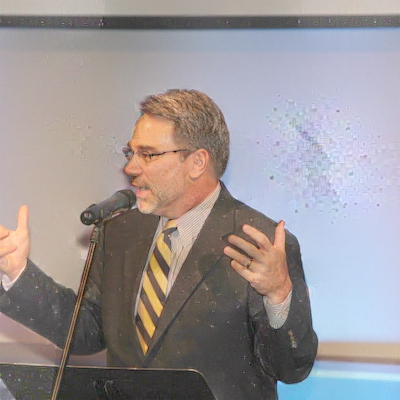}\vspace{\hh}\\
\makebox[\h]{\small{Rel. $L_2$ dist.:}}\hfill
\makebox[\h]{\small{1}}\hfill
\makebox[\h]{\small{16.0}}\hfill
\makebox[\h]{\small{14.0}}\hfill
\makebox[\h]{\small{14.5}}\hfill
\makebox[\h]{\small{3.7}}\vspace{\hhh}\\
\includegraphics[width=\h]{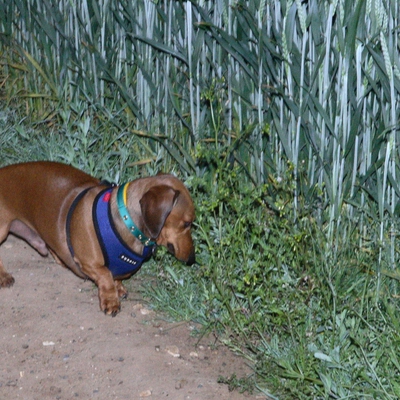}\hfill
\includegraphics[width=\h]{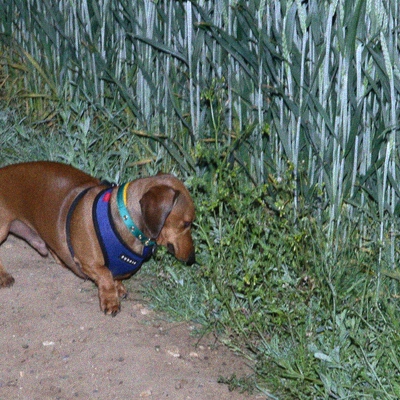}\hfill
\includegraphics[width=\h]{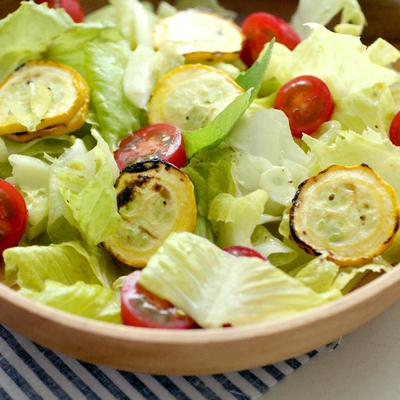}\hfill
\includegraphics[width=\h]{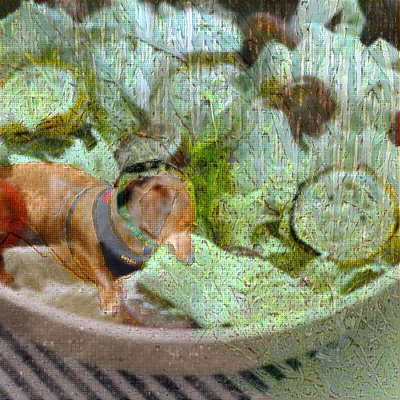}\hfill
\includegraphics[width=\h]{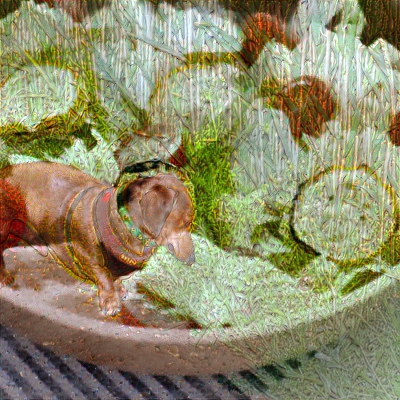}\hfill
\includegraphics[width=\h]{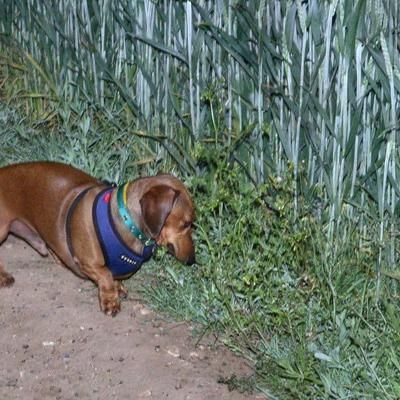}\vspace{\hh}\\
\makebox[\h]{\small{Rel. $L_2$ dist.:}}\hfill
\makebox[\h]{\small{1}}\hfill
\makebox[\h]{\small{13.9}}\hfill
\makebox[\h]{\small{9.2}}\hfill
\makebox[\h]{\small{10.2}}\hfill
\makebox[\h]{\small{1.5}}\vspace{\hhh}\\
\caption{\label{fig:ContradictionA}Attack (A1). 
Both \lpips metrics allow to pull far away from the source towards the target while remaining at the same \lpips distance from the source as the anchor. With the same constraint,  attacks on \ouralgo lie much closer to the source image both visually and by relative $L_2$ distance.}
\end{center}
\end{figure*}
}

\newcommand{\figContradictionBB}{
\begin{figure*}[t]
\renewcommand{\h}{0.12\linewidth}
\renewcommand{\hh}{-1mm}
\renewcommand{\hhh}{1mm}
\makebox[\h][c]{Source $a$}\hfill
\parbox{\h}{\centering Attack on \lpipssqz}\hfill
\parbox{\h}{\centering Attack on \lpipsvgg}\hfill
\parbox{\h}{\centering Attack on \ouralgo (ours)}\,
\makebox[\h][c]{Source $a$}\hfill
\parbox{\h}{\centering Attack on \lpipssqz}\hfill
\parbox{\h}{\centering Attack on \lpipsvgg}\hfill
\parbox{\h}{\centering Attack on \ouralgo (ours)}\,

\includegraphics[width=\h,trim=120 210 210 120,clip]{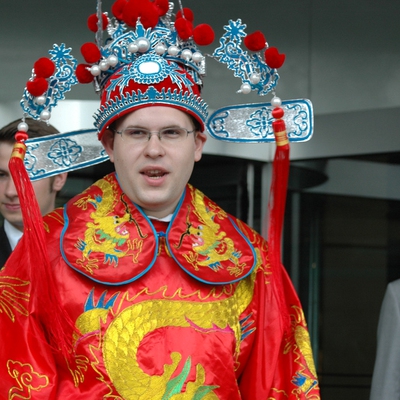}\hfill
\includegraphics[width=\h,trim=120 210 210 120,clip]{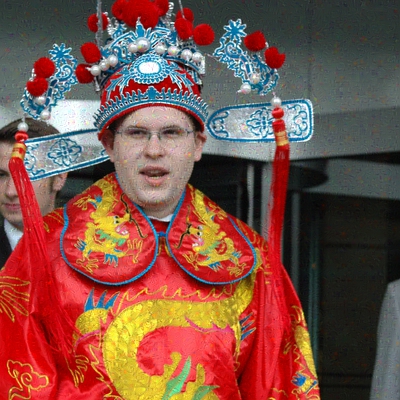}\hfill
\includegraphics[width=\h,trim=120 210 210 120,clip]{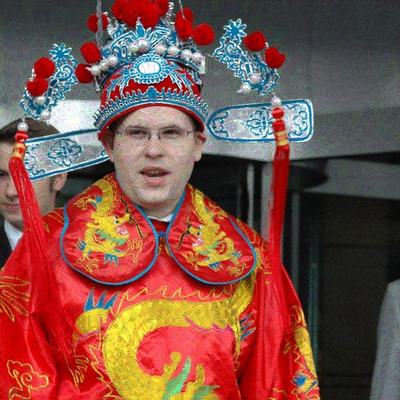}\hfill
\includegraphics[width=\h,trim=120 210 210 120,clip]{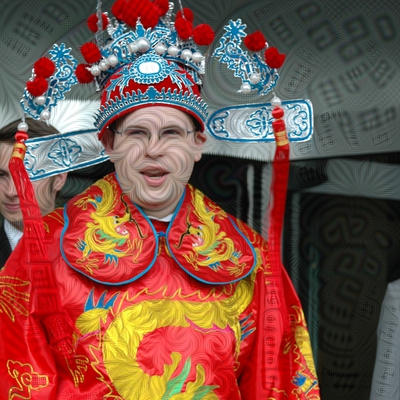}\,
\includegraphics[width=\h,trim=300 280 30 50,clip]{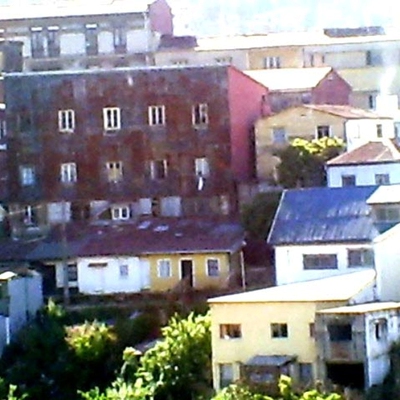}\hfill
\includegraphics[width=\h,trim=300 280 30 50,clip]{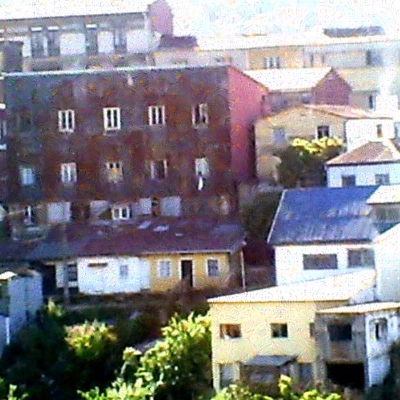}\hfill
\includegraphics[width=\h,trim=300 280 30 50,clip]{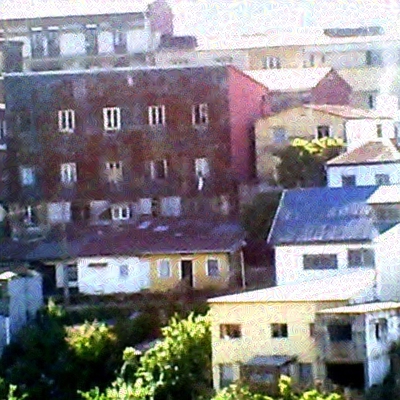}\hfill
\includegraphics[width=\h,trim=300 280 30 50,clip]{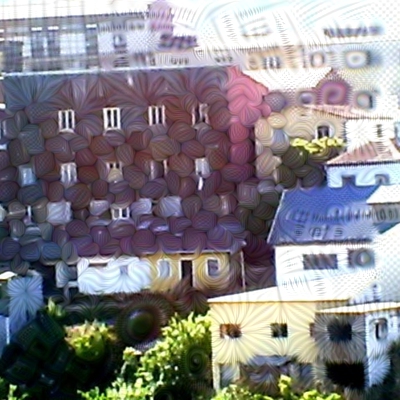}\vspace{\hh}\\
\makebox[\h]{\small{\textsc{(e-)lpips}:}}\hfill
\makebox[\h]{\small{1.36}}\hfill
\makebox[\h]{\small{2.83}}\hfill
\makebox[\h]{\small{0.35}}\,
\makebox[\h]{\small{\textsc{(e-)lpips}:}}\hfill
\makebox[\h]{\small{1.59}}\hfill
\makebox[\h]{\small{2.95}}\hfill
\makebox[\h]{\small{0.58}}\vspace{\hhh}
\caption{\label{fig:ContradictionB}Attack (A2). 
Both \lpips metrics allow the image to be pushed far away in distance by modifications that are small both visually and in $L_2$ sense.  In contrast, the attack is unable to increase the \ouralgo distance nearly as much; furthermore the visual change is much more clearly visible at the same $L_2$ distance, which is desirable. The relative distance reported below the images is normalized such that 1 is the mean distance between the different images in the dataset.}
\end{figure*}
}

\newcommand{\figNoiseBarycenter}{
\begin{figure*}[t]
\begin{center}
\renewcommand{\h}{0.14\linewidth}
\renewcommand{\hh}{1mm}
\renewcommand{\hhh}{11mm}
\parbox{\h}{\centering $x_1$}
\parbox{\h}{\centering $x_2$}
\makebox[6mm]{$\cdots$}%
\parbox{\h}{\centering $x_{10}$}%
\hspace{\hhh}
\parbox{\h}{\centering \ouralgo (ours)}
\parbox{\h}{\centering$L_2$}
\parbox{\h}{\centering\lpips}\\
\raisebox{-.45\height}{\includegraphics[width=\h,trim=220 141 224 123,clip]{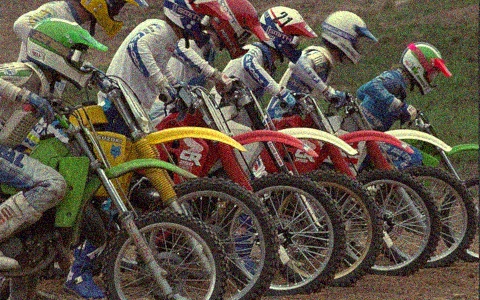}}
\raisebox{-.45\height}{\includegraphics[width=\h,trim=220 141 224 123,clip]{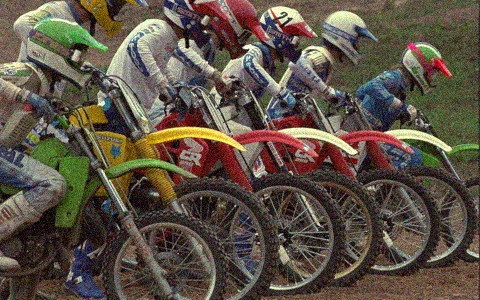}}
\makebox[6mm]{$\cdots$}%
\raisebox{-.45\height}{\includegraphics[width=\h,trim=220 141 224 123,clip]{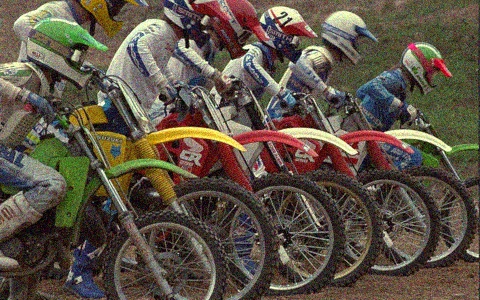}}%
\makebox[\hhh]{$\Longrightarrow$}
\raisebox{-.45\height}{\includegraphics[width=\h,trim=220 141 224 123,clip]{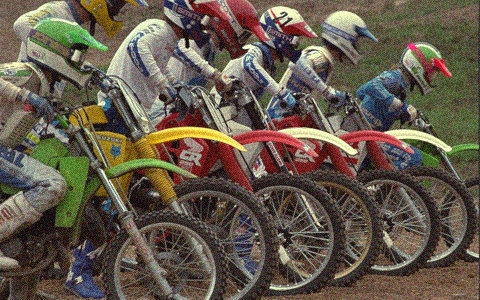}}
\raisebox{-.45\height}{\includegraphics[width=\h,trim=220 141 224 123,clip]{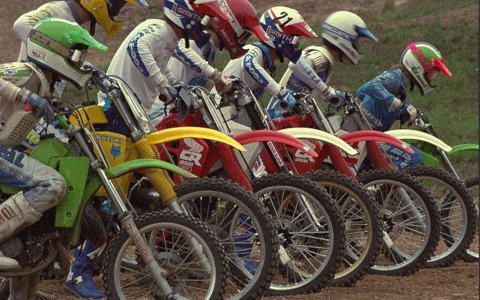}}
\raisebox{-.45\height}{\includegraphics[width=\h,trim=220 141 224 123,clip]{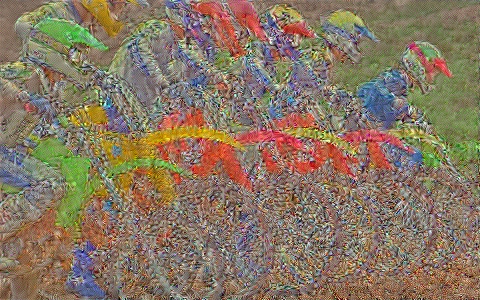}}\vspace{0.5mm}\\
\raisebox{-.45\height}{\includegraphics[width=\h,trim=220 141 224 123,clip]{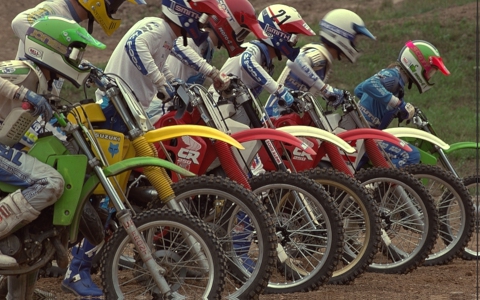}}
\raisebox{-.45\height}{\includegraphics[width=\h,trim=220 141 224 123,clip]{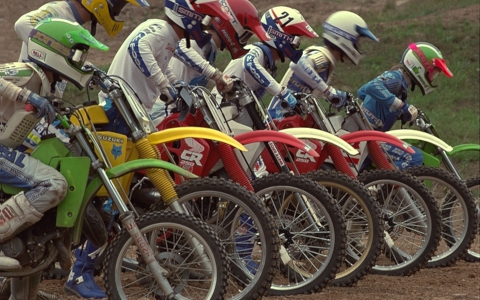}}
\makebox[6mm]{$\cdots$}%
\raisebox{-.45\height}{\includegraphics[width=\h,trim=220 141 224 123,clip]{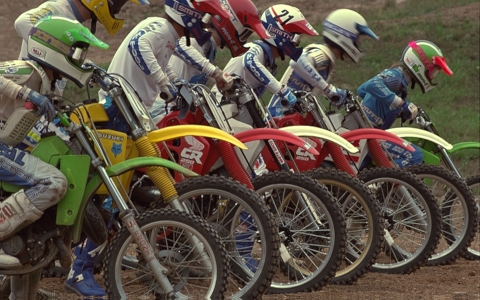}}%
\makebox[\hhh]{$\Longrightarrow$}
\raisebox{-.45\height}{\includegraphics[width=\h,trim=220 141 224 123,clip]{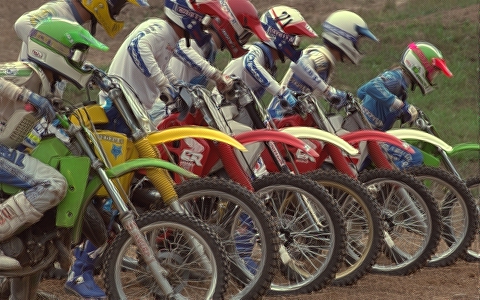}}
\raisebox{-.45\height}{\includegraphics[width=\h,trim=220 141 224 123,clip]{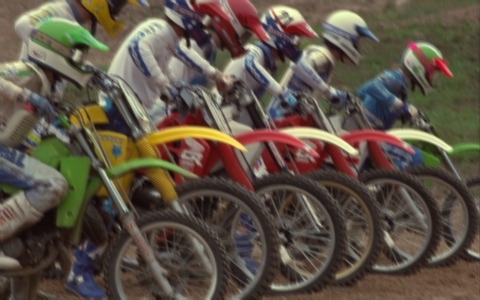}}
\raisebox{-.45\height}{\includegraphics[width=\h,trim=220 141 224 123,clip]{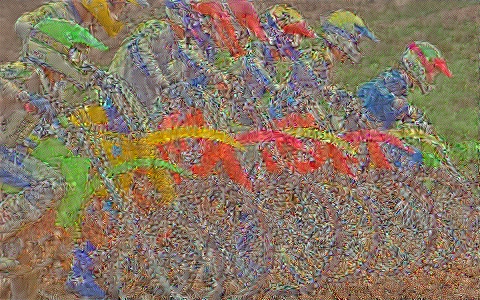}}\\
\makebox[6.75cm]{Input images}\hfill
\makebox[6cm]{Barycenters}
\caption{\label{fig:NoiseBarycenter}Barycenters of similar-looking images $x_1, x_2, \cdots, x_{10}$ under various distance metrics. Top row: 10 noise realizations of the same image. Bottom row: small translations of the same image. The $L_2$ barycenter simply averages the inputs. Unlike $L_2$ and \lpips, the \ouralgo barycenter retains much of the appearance of the input images in both cases.
}
\end{center}
\end{figure*}
}

\newcommand{\figAblation}{
\renewcommand{\h}{0.11\linewidth}
\renewcommand{\hhh}{1mm}
\newcommand{\overlayLabel}[2]{
\begin{tikzpicture}[every node/.style={inner sep=0,outer sep=0}]
\node at (0,0) {\includegraphics[width=\h]{\figdir/rebuttal_ablation1/##1}};
\draw (-5.5mm,6.5mm) node [white] {\small ##2};
\end{tikzpicture}
}
\begin{figure}[t]
\overlayLabel{src_0.jpg}{(a)}\hfill
\overlayLabel{dest_0.jpg}{(b)}\hfill
\overlayLabel{vgg_0.jpg}{(c)}\hfill
\overlayLabel{elpips_plain_0.jpg}{(d)}\hfill
\overlayLabel{elpips_g_0.jpg}{(e)}\hfill
\overlayLabel{elpips_gc_0.jpg}{(f)}\hfill
\overlayLabel{elpips_gcs_0.jpg}{(g)}\hfill
\overlayLabel{elpips_gcsd_0.jpg}{(h)}\\
\makebox[\h]{Source}\hfill
\makebox[\h]{Target}\hfill
\makebox[\h]{\lpipsvgg}\hfill
\makebox[\h]{+all layers}\hfill
\makebox[\h]{+geom.}\hfill
\makebox[\h]{+color}\hfill
\makebox[\h]{+scaling}\hfill
\makebox[\h]{+dropout}  
\caption{\label{fig:Ablation}Success of Attack (A1) against increasingly powerful variants of \ouralgo. The increasing robustness resulting from a richer transformation ensemble is visible as the increasing visual similarity between (a) and attack results (c)-(h). Image (h) corresponds to the full \ouralgo metric.}
\end{figure}
}

\newcommand{\eigenTableBody}{
\begin{tabular}{r||c|c|c} 
\hline
 & \ouralgo & \lpips & ratio ($2\sigma$ bounds) \\ 
\hline
$\lambdamax$                        &   $375$                      &   $4540$         &  $[0.058, 0.146]$   \\
$\lambdamin{}^\dagger$              &   $0.008\,55$                  &   $0.000\,005\,52$   &  $[1090, 2480]$     \\
$\lambdamax / \lambdamin{}^\dagger$ &   $25\,600$                    &   $112\,000\,000$    &  $[0.0002, 0.0005]$ \\
Variance                            &   $11.6$                     &   $422$          &  $[0.031, 0.050]$   \\
Skewness                            &   $24.1$                     &   $87.5$         &  $[0.259, 0.372]$   \\  
Kurtosis                            &   $1\,350$                     &   $13\,600$        &  $[0.105, 0.209]$   \\
\hline
\end{tabular}\vspace{2mm}
}

\newcommand{\eigenTableWrap}{
\setlength\intextsep{0pt}
\begin{wrapfigure}{r}{0.55\textwidth}
\scriptsize
\eigenTableBody
\end{wrapfigure}
}

\usepackage[hidelinks,colorlinks=false,bookmarks=false,hyperindex,pagebackref,citecolor=black]{hyperref}  %

\begin{document}

\title{E-LPIPS: Robust Perceptual Image Similarity via Random Transformation Ensembles}

\author{Markus Kettunen\\
Aalto University
\And
Erik H\"ark\"onen\\
Aalto University
\And
Jaakko Lehtinen\\
Aalto University\\
NVIDIA
}

\maketitle

\begin{abstract}
It has been recently shown that the hidden variables of convolutional neural networks make for an efficient perceptual similarity metric that accurately predicts human judgment on relative image similarity assessment. First, we show that such learned perceptual similarity metrics (LPIPS) are susceptible to adversarial attacks that dramatically contradict human visual similarity judgment. While this is not surprising in light of neural networks' well-known weakness to adversarial perturbations, we proceed to show that self-ensembling with an infinite family of random transformations of the input --- a technique known \emph{not} to render classification networks robust --- is enough to turn the metric robust against attack, while retaining predictive power on human judgments. Finally, we study the geometry imposed by our our novel self-ensembled metric (E-LPIPS) on the space of natural images. We find evidence of ``perceptual convexity'' by showing that convex combinations of similar-looking images retain appearance, and that discrete geodesics yield meaningful frame interpolation and texture morphing, all without explicit correspondences.  
\end{abstract}

\section{Introduction}
\label{sec:intro}

Computational assessment of perceptual image similarity --- how close is an image to another image --- is a fundamental question in vision, graphics, and imaging. For an image similarity metric $d(a, b)$ to be perceptually equivalent to human observations, it clearly suffices that
\begin{enumerate}
    \item[(C1)] $d(a, b)$ is large when human observers perceive a large dissimilarity between $a$ and $b$, and
    \item[(C2)] $d(a, b)$ is small when human observers consider the images $a$ and $b$ similar.
\end{enumerate}

Perceptually motivated image similarity metrics have a long history, with well-known challenges due to dependence on context and high-level image structure \cite{ssim,ms-ssim,Mantiuk:2011:HCV}. Standard vector norms applied to images pixelwise, such as the $L_2$ or $L_1$ distance, are brittle in the sense that many transformations that leave images visually indistinguishable --- shifting by one pixel is a classic example --- can yield arbitrary jumps in distance, indicating a violation of Condition (C2). Other transformations exhibiting similar issues include intensity variations, small rotations, and slight blur. 

In recent years, several authors have made the observation that the hidden layer features in a convolutional neural network (CNN) trained for image classification \cite{Simonyan2014VeryDC,Krizhevsky2014OneWT,SqueezeNet} yields a perceptually meaningful space in which to measure image distance \cite{Berardino2017EigenDistortionsOH}, with applications in, e.g., image generation and restoration \cite{Johnson2016PerceptualLF,Chen2017PhotographicIS,KimLL15} and performance metrics for generative models \cite{Heusel2017}. Inspired by this, Zhang et al. \cite{Zhang_2018_CVPR} recently showed that \emph{hidden CNN activations are indeed a space where distance can strongly correlate with human judgment}, much more so than per-pixel metrics or other prior similarity scores. Their Learned Perceptual Image Path Similarity (\lpips) metric is calibrated to human judgment using data from two-alternative forced choice (2AFC) tests.   

Our message is threefold. First, we will show that \lpips-style image similarity metrics computed as Euclidean distances between the hidden variables of a classifier CNN are brittle: they are easy to break by standard adversarial attack techniques. This is not at all surprising given neural networks' well-known weakness to adversarial perturbations. More precisely, we show that these metrics fulfill neither (C1) nor (C2) by showing how to easily craft image pairs that have a small \lpips distance albeit human observers see the images as completely different, violating (C1), or that have a large \lpips distance, although a human observer is hard pressed to see the difference, violating (C2).

Second, we show that computing the CNN feature distance as an average under an effectively infinite ensemble of random image transformations makes the metric significantly more robust against adversarial attacks, even with the attack crafted with full knowledge of the ensemble defense.
We call the novel metric the Ensembled Learned Perceptual Image Patch Similarity metric, \ouralgo. That self-ensembling through random transformations yields a robust similarity metric might seem surprising at first: similar random transformation ensembles have been previously shown to \emph{not} be sufficient to make classifier networks robust \cite{Athalye2018,Athalye2018Synthesizing}. We attribute the qualitative difference to the fact that attacking \lpips-style metrics requires modification of \emph{all} hidden variables of a CNN, while attacking a classifier network only requires the attacker to change the output layer. 

Finally, we observe that the geometry on the space of natural images induced by the \ouralgo metric has several intriguing properties we call ``perceptual convexity''. In particular, we show by several examples that when computed under our \ouralgo metric, averages (barycenters) of similar-looking images retain a similar appearance even when pixel-wise averages appear very different, and that discrete geodesics between two images yields reasonable frame interpolation and texture morphing --- all without explicit correspondence or optical flow. Furthermore, as a practical consequence of a more perceptually convex metric, we show that \ouralgo yields consistently better results than non-ensembled feature losses when used as a loss function in training image restoration models.

\section{Ensembled Perceptual Similarity}

    We build on \lpips \cite{Zhang_2018_CVPR} that computes image differences in the space of hidden unit activations resulting from the input images when run through the convolutional part of an image classification network such as VGG \cite{Simonyan2014VeryDC}, SqueezeNet \cite{SqueezeNet} or AlexNet \cite{Krizhevsky2014OneWT}. 
    Other, prior uses of ``VGG feature distances'' \cite{Johnson2016PerceptualLF,Chen2017PhotographicIS,KimLL15} differ from \lpips in minor details only.
    
    \lpips first normalizes the feature dimension in all pixels and layers to unit length, scales each feature by a feature-specific weight, and evaluates the square of the $L_2$ distance between these weighted activations. These squared distances are then averaged over the image dimensions and summed over the layers, and this provides the final image distance metric:
\begin{equation}
    d_\lpips(x,y) =  \sum_l{\frac{1}{H_l W_l}} \sum_{i,j} \Vert w_l \odot (\hat{x}^l_{ij} - \hat{y}^l_{ij}) \Vert^2_2. \label{eq:lpips}
\end{equation}
    Here, $\hat{x}^l_{ij}$ and $\hat{y}^l_{ij}$ denote the normalized feature vectors at layer $l$ and pixel $(i, j)$, $w_l$ contains weights for each of the features in layer $l$, and $\odot$ multiplies the feature vectors at each pixel by the feature weights.
    The weights $w_l$ are optimized such that the metric best agrees with human judgment derived from two-alternative forced choice (2AFC) test results. 
    
\subsection{E-LPIPS: Ensembled LPIPS}
    Below (Section~\ref{sec:contradictions}), we prove that \lpips fulfills neither (C1) nor (C2) by describing simple adversarial attacks that violate both conditions. The reader may consult Figures~\ref{fig:ContradictionA} and \ref{fig:ContradictionB} before proceeding.
    With hopes of increasing its robustness, we alter \lpips in four ways:
    \begin{enumerate}
        \item We apply a randomized geometric and color transformation to the input images before feeding them to the network, and take the expected \lpips distance over realizations.  As the model is not limited to a single glimpse of the input, but has access to an essentially infinite number of different versions, it is harder to fool.
        \item While \lpips only considers the features in the last layer of each resolution, we compute distances over \emph{all} convolutional layers. In particular, while \lpips is blind to the input layer, considering all layers means Equation~\eqref{eq:lpips} includes the $L_2$ distance of the input color tones, with a learned weight $w_0$. 
        \item We make the network stochastic by applying dropout with keep probability $0.99$ on all layers. Matching intuition, this makes the metric consistently more robust. 
        \item We replace the image classification networks' max pooling layers with average pooling. As observed by several prior authors \cite{Gatys2015,Henaff2016}, max pooling distracts gradient flow, making optimization harder, and leaves more blind spots to be exploited since many of the network activations can be freely altered without the change propagating forward.
    \end{enumerate}

    For a single evaluation of our method, we first transform both input images with the same random combination of simple transformations.
    Our transform set includes translations, flipping, mirroring, permutation of color channels, scalar multiplication (brightness change), and downscaling, all with randomly chosen parameters. Furthermore, the simple random transformations are combined randomly, making the effective size of the ensemble very large. This is in stark contrast to an ensemble over a fixed, small number of separate transformations.  Algorithms \ref{alg:SampleTransform} and \ref{alg:ApplyTransform} detail the construction and application of the transformations.

    We now define \ouralgo as the expectation
    \begin{equation}
        d_\ouralgo(x,y) = \E[d_\lpips(T(x), T(y))]
    \end{equation}
    where the expectation is taken over the stochastic combinations of transformations $T$ described above, and the classical $d_\lpips$ distance evaluated with our network with dropout, average pooling, and weights applied to all layers, including the input layer.

    For numerical optimization using the metric, e.g., training neural networks, we find that a single random sample per iteration is generally sufficient, but multiple samples may sometimes result in faster convergence. For more precise distance evaluations we recommend averaging $N \geq 300$ evaluations, depending on the required precision.
    For comparing distances of two images from a third image, 
    we recommend using the same transformations and dropout variables for all images. A single evaluation of \ouralgo without gradients takes on average about 10\% longer than \lpipsvgg, and 20\% longer with gradients.

\begin{minipage}[t]{0.5\textwidth}
\begin{algorithm}[H]
    \SetKwInOut{Input}{Input}
    \SetKwInOut{Output}{Output}

    \Output{Random transformation parameters}
    dx, dy = random.int$(0\hdots 7)$\;
    flip\_x, flip\_y, swap\_xy = random.int() \% 2\;
    col\_perm = random.permute$(0, 1, 2)$\;
    sr, sg, sb = random.uniform$(0.2, 1.0)$\;
    scale = random.int$(1\hdots 8)$ with $p(i) \propto \frac{1}{i^2}$\;
    scale\_dx, scale\_dy = random.int$(0\ \hdots\ \text{scale - 1})$
    \caption{Sampling a transformation}
    \label{alg:SampleTransform}
\end{algorithm}
\end{minipage}\hfill
\begin{minipage}[t]{0.46\textwidth}
\begin{algorithm}[H]
    \SetKwInOut{Input}{Input}
    \SetKwInOut{Output}{Output}

    \Input{Image $X$, transformation parameters}
    \Output{Transformed image}
    X = mirrorPad(X, scale\_dx, scale\_dy, 0, 0)\;
    X = downscaleBox(X, 1/scale)\;
    X = mirrorPad(X, dx, dy, $7-\text{dx}$, $7-\text{dy}$)\;
    \lIf{flip\_x}{X = flipX(X)}
    \lIf{flip\_y}{X = flipY(X)}
    \lIf{swap\_xy}{X = swapXY(X)}
    X = permuteColorChannels(X, col\_perm)\;
    X = X * (sr, sg, sb)\;

    \caption{Applying a transformation}
    \label{alg:ApplyTransform}
\end{algorithm}
\end{minipage}

\section{Experiments}
\label{sec:experiments}
\figContradictionAA

This section studies the properties of \ouralgo in comparison to \lpips and simpler metrics in various scenarios. We use the VGG-16 version as basis for \ouralgo in all results, but compare to both VGG and Squeezenet -based \lpips, the best models reported by Zhang et al.~\cite{Zhang_2018_CVPR}. All models are implemented in TensorFlow, and all training and other optimization is performed on NVIDIA V100 GPUs.
We perform all optimizations with Adam \cite{kingma2014adam}, propagating gradients the random input transformations. When attacking the metrics, this directly implements the \emph{Expectation over Transformations} (EoT) attack previously used to break self-ensembling in image classifiers~\cite{Carlini2017TowardsET,Athalye2018}. 

\subsection{Adversarial Examples}
\label{sec:contradictions}

\paragraph{Attack (A1), violates (C1).} We construct image pairs that humans perceive dissimilar but \lpips considers close. To attack metric $d(\cdot, \cdot)$, we select a source image $a$, constrain $d(a, x)$ to be small, and optimize $x$ towards a distant target image $b$ in the $L_2$ sense by solving
\begin{equation}
    \argmin_x \|x - b\|^2_2 \quad \text{subject to} \quad d(a, x) \le \varepsilon.
\end{equation}
with the anchor distance $\varepsilon$ chosen as the distance between $a$ and a slightly noisy version of $a$. We perform the attack on 30 images from the OpenImages~\cite{OpenImages} dataset.
Figure~\ref{fig:ContradictionA} shows representative results. Both \lpips metrics succumb badly: the result shows that the $\varepsilon$-\lpips-ball around the source image contains images that are wildly different from it. The \ouralgo images consistently stay much closer to the input image, as desirable.
 On average, \lpips lets the attack pull the images five times farther from the input image in $L_2$ distance than \ouralgo: $16.2 \pm 0.8$ times $\varepsilon$ for \lpipsvgg, $15.2 \pm 0.7$ for \lpipssqz, and $3.0 \pm 0.2$ for \ouralgo. See the supplemental material for more examples.

\paragraph{Attack (A2), violates (C2).} In the opposite direction, we construct pairs of images  that are similar to the human eye (proxied by a small $L_2$ distance), but far apart in the \lpips sense by solving
\begin{equation}
    \argmax_x d(a, x) \quad \text{subject to} \quad \|a-x\|^2_2 \le \varepsilon.
\end{equation}
The other details of the attack are the same as in (A1).
Figure~\ref{fig:ContradictionB} shows representative results. Both \lpips metrics allow a very large increase distance while staying within a small $L_2$-ball around the input: $1.74 \pm 0.07$ times the mean distance between the images in the dataset for \lpipssqz, on average, and $3.12 \pm 0.08$ for \lpipsvgg. In contrast, \ouralgo gives in much less: $0.63 \pm 0.06$.

For \lpips, the attack images are visually highly similar to the source, despite the large increase in the distance, indicating violation of (C2); 
in contrast, the image of maximal \ouralgo distance within the $\varepsilon$-$L_2$-ball is consistently visibly more different from the source image. This is desirable. The supplemental material contains more examples.

\figContradictionBB

\subsection{Analysis}
Together, the attack results suggest that \ouralgo fulfills conditions (C1) and (C2) better than the non-ensembled feature metrics. We now proceed to study the reasons for this result.

\paragraph{Are all the transformations necessary?} We study the necessity of the various random transformations in our ensemble by an ablation. We repeat Attack (A1) on an image pair, adding transformations to the ensemble one-by-one, with dropout being added last. Figure~\ref{fig:Ablation} shows representative results: clearly, each individual transformation ``plugs more holes'', making the attack less and less successful. On a test set of 30 images, the mean distance of \lpipsvgg from the clean image is $16.3 \pm 0.8$ times the anchor distance. The effectiveness of the adversarial attack drops as we add all layers and average pooling ($14.5 \pm 0.7$), translations, rotations and swaps ($10.8 \pm 0.7$), color transformations ($6.3 \pm 0.5$), multiple scale levels ($4.6 \pm 0.4$), and finally dropout ($3.0 \pm 0.2$), reaching a five-fold robustness improvement in terms of the $L_2$ distance.

The supplemental material contains another ablation study that highlights the necessity of using an effectively infinite ensemble of random transformations, as opposed to a fixed ensemble.

\figAblation

\paragraph{Sensitivity analysis.}
Above, we found large $L_2$ perturbations that increased the \lpips distance only little, and small $L_2$ perturbations that greatly increased the \lpips distance. The results are systematic: attacks succeed easily against \lpips, and much less so against \ouralgo. This suggests severe ill-conditioning of the \lpips metric. 
We now probe the metrics' local behavior in order to see if it is consistent with this hypothesis.
Specifically, we study the behavior of $f_{x_0}(v) = d(x_0, x_0 + v)$ around a fixed image $x_0$ for small $v$ through its 2nd-order Taylor expansion. As $v=0$ is the minimum of $f(v)$, the first-order term vanishes, and the expansion equals $v^T H v / 2$, where $H=\partial^2 f/\partial v^2$ is the Hessian. Note that $\dim v=W\times H\times 3$ for an RGB image of size $W\times H$. 

\eigenTableWrap
As the Hessian is symmetric and positive semidefinite, its distance-scaling properties are characterized by its eigenvalue spectrum. 
While the full spectrum is intractable, we compute the extremal eigenvalues by power iteration, as well as approximate their mean, variance, skewness and kurtosis by sampling. Details can be found in the supplemental material. The adjacent table shows means of these descriptors computed over 10 different anchor images $x_0$ of size $400 \times 400$. For scale-invariance between the metrics, we normalize all values with the corresponding mean eigenvalues. The smallest eigenvalues of \ouralgo are challenging to compute due to the metric's stochastic nature. We thus evaluate the results marked with $\dagger$ in lower resolution ($64 \times 64$), and use a weakened version of \ouralgo with no dropout and with a fixed ensemble of 256 input transformations. We notice a clear trend of improved results when the ensemble size is increased, suggesting still better conditioning for full \ouralgo. 

The maximal eigenvalues of \lpips are consistently an order of magnitude larger than \ouralgo', the minimal eigenvalues are consistently three orders of magnitude smaller, and \lpips' condition number is three orders of magnitude larger on average. These numbers indicate that \lpips is much more ill-conditioned.
Moreover, the distribution of \lpips' eigenvalues is more positively skew --- indicating that more eigenvalues are clustered towards zero from the mean and consequently more directions where the metric is blind to changes --- as well as more kurtotic, implying more concentration at the extremal ends. These findings are consistent with the large-scale behavior observed earlier. 

The supplemental material contains videos demonstrating random perturbations in the linear space spanned by the 16 largest eigenvectors of both \lpips and \ouralgo around the central image $x_0$. Though informal, the perturbations sampled from \ouralgo eigenvectors appear more natural and consistent with the image structure.

\paragraph{Correlation with Human Judgment.}
\label{sec:AFC}%
We now study the power of \ouralgo in predicting human assessment of image similarity, the original goal of the \lpips metric that ours is based on.
We use the data of, and directly follow the evaluation process presented by Zhang et al. \cite{Zhang_2018_CVPR}. We train the feature weights to produce distances $d(A, C)$ and $d(B, C)$ which are fed to a small two-hidden-layer MLP that predicts the ratio of human choices of A and B. We enable input transformations only for testing after the weights have been trained, and keep the weights of the underlying VGG network fixed.

The resulting accuracy in predicting human answers in the two-alternative forced choice test (2AFC) is 69.16\% for \ouralgo, falling directly between the VGG (68.65\%) and SqueezeNet (70.07\%) versions of the non-ensembled \lpips. This indicates that predictive power does not suffer as a result of the ensembling, i.e., additional robustness and more intuitive geometry over images come at no cost. To put the numbers in perspective, Zhang et al. report a mean human score of 73.9\%\footnote{The score is not 100\%, as humans have difficulty predicting how other humans perceive image similarity.}, while simple metrics such as pixel-wise $L_2$ distance and SSIM achieve the score of 63\%. The supplemental material contains a table with an accuracy breakdown over different classes of image corruptions.

\paragraph{Performance as a Loss Function.}
\label{sec:trainingcnns}
Using an image similarity metric as a loss function in training an image generation or restoration model is a potential adversarial situation: if the metric cannot reliably distinguish between perceptual similarity and dissimilarity, the optimizer may drive the model to produce nonsensical results.
As a case study, we train a 10M parameter standard convolutional U-net with skip connections \cite{Ronneberger2015} for removing additive Gaussian noise from photographs using different loss functions, as well as for 4-fold single-image super-resolution. (We train super-resolution in a supervised fashion \cite{Johnson2016PerceptualLF} as a case study, aware that state-of-the-art techniques utilize more sophisticated models \cite{Dahl2017,Ledig2017PhotoRealisticSI}.)

Although the differences are not large on average, we find that \ouralgo consistently results in somewhat better results than plain \lpips, both numerically and by visual inspection. In particular, the non-ensembled \lpips metric sometimes falls in strange local minima on some network architectures (e.g. transposed convolutions, see supplemental). This is never observed with \ouralgo. Despite the increased robustness and somewhat better results, there is no difference in wall-clock training speed between \lpips and \ouralgo. Visual and numerical results can be found in the supplemental material.

\subsection{Geometry of Natural Images}
\figNoiseBarycenter

\paragraph{Barycenters of similar-looking images.}
From a perceptual point of view, it would seem reasonable that an average of two or more images that look the same should still look the same. To study the properties of averages of multiple images taken under \ouralgo, we first compute barycenters, $\argmin_x \sum_i d(x, n_i)^2$, over several independent realizations $\{n_i\}$ of images transformed by the same random process, by direct numerical optimization over pixels.

As examples, we study i.i.d. additive Gaussian noise and small random offsets in Figure~\ref{fig:NoiseBarycenter}. The $L_2$ barycenter is simply the pixelwise average of the inputs, and appears quite different from the individual input images: averaging zero-mean noise cancels it out, and averaging shifted images results in a blur. Despite having converged into a local optimum, the non-ensembled \lpips barycenters deviate visually even further, providing another view into the metric's large null space. In contrast, the appearance of the \ouralgo barycenters quite closely matches the individual realizations. This suggests a property we call \emph{perceptual convexity}: the barycenter --- which can be seen as a convex combination --- of a set of visually similar images is itself visually similar to the others. The supplemental material contains more examples, and a study of pairwise averages.

\paragraph{Discrete geodesics.}
\figGeodesic
A discrete geodesic between images $a$ and $b$ is a sequence of images that minimizes the total pairwise squared distance between adjacent frames \cite{Henaff2016}:
\begin{equation}
\argmin_{x_1, \hdots, x_{N-1}} \sum_{i=0}^{N-1} d(x_i, x_{i+1})^2, \quad \text{with } x_0 = a, \;x_N = b. \label{eq:geodesic}
\end{equation}
The metric $d(\cdot, \cdot)$ directly determines the visual properties of the solution, as the free images $x_1, \hdots, x_{N-1}$ affect the optimization objective only through it. The $L_2$ geodesic is a pixelwise linear cross-fade.
We numerically compute discrete geodesics of 8 in-between frames for the \ouralgo, \lpips, and $L_2$ metrics by initializing the free frames to random noise, and simultaneously optimize all of them  using Adam. See the supplemental material for details on the optimization procedure. 

As shown in Figure~\ref{fig:Geodesic}, geodesic interpolation between adjacent video frames appears to create meaningful frame interpolations in the \ouralgo metric: instead of fading in and out like the $L_2$ crossfade, many image features translate and rotate as if reprojected along optical flow. Note that no correspondence or explicit reprojection is performed; all effects emerge from the metric. In addition to frame interpolation, the geodesics result in interesting fusions between different textures. Some evidence of similar behavior is seen in \lpips geodesics, but the image fusion is generally of poorer quality, and translation and rotation are not modeled as naturally. The reader is encouraged to view the videos in the supplemental material. Due to the difficulty of the optimization problem, we consider these results an existence proof that warrants detailed further study.

\section{Related Work}

\paragraph{Adversarial attacks.}
The output of deep neural networks can often be altered drastically with small adversarial perturbations of the input found by gradient-based optimization \cite{IntriguingProperties,Carlini2017TowardsET, Goodfellow2014ExplainingAH, MoosaviDezfooli2016DeepFoolAS, Bhagoji2016DimensionalityRA, Papernot2016DistillationAA}. 
Random self-ensembling has been suggested before as a defense against attacks on classifiers. Li et al.~\cite{Liu2017TowardsRN} inject noise layers into an image classifier. Xie et al.~\cite{XieMitigatingAE2017} and Athalye et al.~\cite{Athalye2018Synthesizing} apply random transformations to the input. These defenses succumb to the Expectation over Transformations (EoT) attack \cite{Athalye2018} --- random transformations are not strong enough to robustify a classifier.

\paragraph{Properties of image metrics.}
Weaknesses in classical image similarity metrics have been previously demonstrated in the same fashion as our attacks
\cite{Wang2009MeanSE, ssim}. We are, however, not aware of prior successful attacks against state-of-the-art neural perceptual similarity metrics. Hessian eigenanalysis of deep convolutional features has been previously employed for studying perturbations of minimal and maximal discriminability on different levels of VGG \cite{Berardino2017EigenDistortionsOH}.

Prior study has hinted toward similar interesting properties of the discrete geodesics of the ``VGG feature'' image metric: Hénaff and Simoncelli \cite{Henaff2016} find evidence of linearization of  effects such as small translation and rotation, but only when aided with additional projections into pixel-space geodesics. In contrast, our geodesics are computed by optimizing only the \ouralgo and \lpips metrics. This, together with the large \lpips null space evident in the barycenter results (Figure~\ref{fig:NoiseBarycenter}), suggests that \ouralgo imposes a more robust geometric structure on the space of images.

\section{Conclusion}
By constructing adversarial examples that demonstrate high perceptual non-uniformity in \lpips \cite{Zhang_2018_CVPR}, we showed that image similarity metrics based on convolutional classifier feature distances exhibit susceptibility to white-box gradient-based perturbation attacks, much like image classifiers. While this is not surprising, the observation explains minor practical robustness issues faced when using such metrics as optimization targets without extra regularization.

We further extended the fragile \lpips metric into an effectively infinite self-ensemble through applying random, simple image transformations and dropout. The resulting \ouralgo image similarity metric ('E' for ensemble) is much more resilient to various direct attacks mounted with full knowledge of the ensemble. This presents an interesting contrast between perceptual feature differences and image classifiers: it is known that similar defenses do not suffice against similar attacks \cite{Athalye2018}. Our preliminary studies indicate the difference can be explained by the much larger dimensionality of the hidden convolutional features --- attacking the similarity metric requires modification of \emph{all} convolutional features, as opposed to just the output layer. While supporting evidence is found from local eigenvalue analysis, we stress that we do not assert theoretical robustness guarantees without further study. 
Still, the observed higher practical resilience to attack is visible as increased robustness when the similarity metric is used as a loss function in optimization. Moreover, the new metric remains a good predictor of human judgment of image similarity, like \lpips.

By computing barycenters of image sets and discrete geodesics between image pairs, we found evidence of ``perceptual convexity'' in the \ouralgo metric: averages taken under it retain a much closer appearance to the images being averaged than under \lpips or $L_2$, and geodesic sequences interestingly linearize effects such as small translation and rotation, without explicit correspondence or optical flow computation. Both findings require systematic further study.

The supplemental material and source code are available at \href{https://github.com/mkettune/elpips/}{\tt github.com/mkettune/elpips}.

\section*{Acknowledgments}
We thank Pauli Kemppinen, Fr\'edo Durand, Miika Aittala, Sylvain Paris, Alexei Efros, Richard Zhang, Taesung Park, Tero Karras, Samuli Laine, Timo Aila, and Antti Tarvainen for in-depth discussions; and Seyoung Park for helping with the TensorFlow port of LPIPS. We acknowledge the computational resources provided by the Aalto Science-IT project.

{
\bibliographystyle{ieee}
\bibliography{000_vggens,001_adversarial}

\begin{thebibliography}{10}\itemsep=-1pt

\bibitem{Athalye2018}
A.~Athalye, N.~Carlini, and D.~A. Wagner.
\newblock Obfuscated gradients give a false sense of security: Circumventing
  defenses to adversarial examples.
\newblock In {\em Proc. 35th International Conference on Machine Learning
  (ICML)}, volume~80, pages 274--283. PMLR, 2018.

\bibitem{Athalye2018Synthesizing}
A.~Athalye, L.~Engstrom, A.~Ilyas, and K.~Kwok.
\newblock Synthesizing robust adversarial examples.
\newblock In {\em Proc. 35th International Conference on Machine Learning
  (ICML)}, volume~80, pages 284--293. PMLR, 2018.

\bibitem{Berardino2017EigenDistortionsOH}
A.~Berardino, J.~Ball{\'e}, V.~Laparra, and E.~P. Simoncelli.
\newblock Eigen-distortions of hierarchical representations.
\newblock In {\em Proc. Neural Information Processing Systems (NeurIPS)}, 2017.

\bibitem{Bhagoji2016DimensionalityRA}
A.~N. Bhagoji, D.~Cullina, and P.~Mittal.
\newblock Dimensionality reduction as a defense against evasion attacks on
  machine learning classifiers.
\newblock {\em CoRR}, abs/1704.02654, 2016.

\bibitem{Carlini2017TowardsET}
N.~Carlini and D.~A. Wagner.
\newblock Towards evaluating the robustness of neural networks.
\newblock {\em 2017 IEEE Symposium on Security and Privacy (SP)}, pages 39--57,
  2017.

\bibitem{Chen2017PhotographicIS}
Q.~Chen and V.~Koltun.
\newblock Photographic image synthesis with cascaded refinement networks.
\newblock {\em 2017 IEEE International Conference on Computer Vision (ICCV)},
  pages 1520--1529, 2017.

\bibitem{Dahl2017}
R.~Dahl, M.~Norouzi, and J.~Shlens.
\newblock Pixel recursive super resolution.
\newblock In {\em 2017 IEEE International Conference on Computer Vision
  (ICCV)}, pages 5449--5458, Oct 2017.

\bibitem{Gatys2015}
L.~Gatys, A.~S. Ecker, and M.~Bethge.
\newblock Texture synthesis using convolutional neural networks.
\newblock In C.~Cortes, N.~D. Lawrence, D.~D. Lee, M.~Sugiyama, and R.~Garnett,
  editors, {\em Proc. Neural Information Processing Systems (NeurIPS)}, pages
  262--270. Curran Associates, Inc., 2015.

\bibitem{Goodfellow2014ExplainingAH}
I.~J. Goodfellow, J.~Shlens, and C.~Szegedy.
\newblock Explaining and harnessing adversarial examples.
\newblock {\em CoRR}, abs/1412.6572, 2014.

\bibitem{Henaff2016}
O.~J. Henaff and E.~P. Simoncelli.
\newblock Geodesics of learned representations.
\newblock In {\em Proc. International Conference on Learning Representations
  (ICLR)}, 2016.

\bibitem{Heusel2017}
M.~Heusel, H.~Ramsauer, T.~Unterthiner, B.~Nessler, and S.~Hochreiter.
\newblock Gans trained by a two time-scale update rule converge to a local nash
  equilibrium.
\newblock In I.~Guyon, U.~V. Luxburg, S.~Bengio, H.~Wallach, R.~Fergus,
  S.~Vishwanathan, and R.~Garnett, editors, {\em Proc. Neural Information
  Processing Systems (NeurIPS)}, pages 6626--6637. Curran Associates, Inc.,
  2017.

\bibitem{SqueezeNet}
F.~N. Iandola, S.~Han, M.~W. Moskewicz, K.~Ashraf, W.~J. Dally, and K.~Keutzer.
\newblock Squeezenet: Alexnet-level accuracy with 50x fewer parameters and
  $<$0.5mb model size.
\newblock {\em arXiv:1602.07360}, 2016.

\bibitem{Johnson2016PerceptualLF}
J.~Johnson, A.~Alahi, and L.~Fei-Fei.
\newblock Perceptual losses for real-time style transfer and super-resolution.
\newblock In {\em ECCV}, 2016.

\bibitem{KimLL15}
J.~Kim, J.~K. Lee, and K.~M. Lee.
\newblock Accurate image super-resolution using very deep convolutional
  networks.
\newblock {\em CoRR}, abs/1511.04587, 2015.

\bibitem{kingma2014adam}
D.~P. Kingma and J.~Ba.
\newblock Adam: A method for stochastic optimization, 2014.

\bibitem{Krizhevsky2014OneWT}
A.~Krizhevsky.
\newblock One weird trick for parallelizing convolutional neural networks.
\newblock {\em CoRR}, abs/1404.5997, 2014.

\bibitem{OpenImages}
A.~Kuznetsova, H.~Rom, N.~Alldrin, J.~Uijlings, I.~Krasin, J.~Pont-Tuset,
  S.~Kamali, S.~Popov, M.~Malloci, T.~Duerig, and V.~Ferrari.
\newblock The open images dataset v4: Unified image classification, object
  detection, and visual relationship detection at scale.
\newblock {\em arXiv:1811.00982}, 2018.

\bibitem{Ledig2017PhotoRealisticSI}
C.~Ledig, L.~Theis, F.~Huszar, J.~Caballero, A.~P. Aitken, A.~Tejani, J.~Totz,
  Z.~Wang, and W.~Shi.
\newblock Photo-realistic single image super-resolution using a generative
  adversarial network.
\newblock {\em 2017 IEEE Conference on Computer Vision and Pattern Recognition
  (CVPR)}, pages 105--114, 2017.

\bibitem{Liu2017TowardsRN}
X.~Liu, M.~Cheng, H.~Zhang, and C.-J. Hsieh.
\newblock Towards robust neural networks via random self-ensemble.
\newblock In {\em Proc. European Conference on Computer Vision (ECCV)}, 2018.

\bibitem{Mantiuk:2011:HCV}
R.~Mantiuk, K.~J. Kim, A.~G. Rempel, and W.~Heidrich.
\newblock Hdr-vdp-2: A calibrated visual metric for visibility and quality
  predictions in all luminance conditions.
\newblock {\em ACM Trans. Graph.}, 30(4):40:1--40:14, July 2011.

\bibitem{MoosaviDezfooli2016DeepFoolAS}
S.-M. Moosavi-Dezfooli, A.~Fawzi, and P.~Frossard.
\newblock Deepfool: A simple and accurate method to fool deep neural networks.
\newblock {\em 2016 IEEE Conference on Computer Vision and Pattern Recognition
  (CVPR)}, pages 2574--2582, 2016.

\bibitem{Papernot2016DistillationAA}
N.~Papernot, P.~D. McDaniel, X.~Wu, S.~Jha, and A.~Swami.
\newblock Distillation as a defense to adversarial perturbations against deep
  neural networks.
\newblock {\em 2016 IEEE Symposium on Security and Privacy (SP)}, pages
  582--597, 2016.

\bibitem{Ronneberger2015}
O.~Ronneberger, P.~Fischer, and T.~Brox.
\newblock U-net: Convolutional networks for biomedical image segmentation.
\newblock {\em MICCAI}, 9351:234--241, 2015.

\bibitem{Simonyan2014VeryDC}
K.~Simonyan and A.~Zisserman.
\newblock Very deep convolutional networks for large-scale image recognition.
\newblock {\em CoRR}, abs/1409.1556, 2014.

\bibitem{IntriguingProperties}
C.~Szegedy, W.~Zaremba, I.~Sutskever, J.~Bruna, D.~Erhan, I.~Goodfellow, and
  R.~Fergus.
\newblock Intriguing properties of neural networks.
\newblock In {\em International Conference on Learning Representations}, 2014.

\bibitem{Wang2009MeanSE}
Z.~Wang and A.~C. Bovik.
\newblock Mean squared error: Love it or leave it? a new look at signal
  fidelity measures.
\newblock {\em IEEE Signal Processing Magazine}, 26:98--117, 2009.

\bibitem{ssim}
Z.~Wang, A.~C. Bovik, H.~R. Sheikh, and E.~P. Simoncelli.
\newblock Image quality assessment: from error visibility to structural
  similarity.
\newblock {\em IEEE Transactions on Image Processing}, 13(4):600--612, April
  2004.

\bibitem{ms-ssim}
Z.~Wang, E.~P. Simoncelli, and A.~C. Bovik.
\newblock Multiscale structural similarity for image quality assessment.
\newblock In {\em The Thrity-Seventh Asilomar Conference on Signals, Systems
  Computers, 2003}, volume~2, pages 1398--1402 Vol.2, Nov 2003.

\bibitem{XieMitigatingAE2017}
C.~Xie, J.~Wang, Z.~Zhang, Z.~Ren, and A.~L. Yuille.
\newblock Mitigating adversarial effects through randomization.
\newblock {\em CoRR}, abs/1711.01991, 2017.

\bibitem{Zhang_2018_CVPR}
R.~Zhang, P.~Isola, A.~A. Efros, E.~Shechtman, and O.~Wang.
\newblock The unreasonable effectiveness of deep features as a perceptual
  metric.
\newblock In {\em The IEEE Conference on Computer Vision and Pattern
  Recognition (CVPR)}, June 2018.

\end{thebibliography}
}

\end{document}